# A Soft-Rigid Hybrid Gripper with Lateral Compliance and Dexterous In-hand Manipulation

Wenpei Zhu[1,2,3#], Chenghua Lu[4#], Qule Zheng[1,2,3], Zhonggui Fang[1,2,3], Haichuan Che[1,2,3],
Kailuan Tang[1,2,3], Mingchao Zhu[4], Sicong Liu[1,2,3] And Zheng Wang[1,2,3*], *Senior Member*, *IEEE*

*Abstract*—Soft grippers are receiving growing attention due to their compliance-based interactive safety and dexterity. Hybrid gripper (soft actuators enhanced by rigid constraints) is a new trend in soft gripper design. With right structural components actuated by soft actuators, they could achieve excellent grasping adaptability and payload, while also being easy to model and control with conventional kinematics. However, existing works were mostly focused on achieving superior payload and perception with simple planar workspaces, resulting in far less dexterity compared with conventional grippers. In this work, we took inspiration from the human Metacarpophalangeal (MCP) joint and proposed a new hybrid gripper design with 8 independent muscles. It was shown that adding the MCP complexity was critical in enabling a range of novel features in the hybrid gripper, including in-hand manipulation, lateral passive compliance, as well as new control modes. A prototype gripper was fabricated and tested on our proprietary dual-arm robot platform with vision guided grasping. With very lightweight pneumatic bellows soft actuators, the gripper could grasp objects over 25 times its own weight with lateral compliance. Using the dual-arm platform, highly anthropomorphic dexterous manipulations were demonstrated using two hybrid grippers, from Tug-of-war on a rigid rod, to passing a soft towel between two grippers using in-hand manipulation. Matching with the novel features and performance specifications of the proposed hybrid gripper, the underlying modeling, actuation, control, and experimental validation details were also presented, offering a promising approach to achieving enhanced dexterity, strength, and compliance in robotic grippers.

## I. Introduction

GRIPPER is the key component for robots to interact with the environment [1, 2]. Design of gripper meets more challenges while robots leave factory and perform activities of daily life (ADL) in domestic environments [3, 4]. ADL involve challenges including object delicacy, environmental uncertainty, and constrained object position beyond the dexterous workspace. Although motor-driven grippers have high precision and large output forces [5, 6], their limited structural compliance [7, 8] requires sensory feedback handling irregular objects under unstructured environments [9-11]. Underactuated grippers use passive structural compliance to enhance environmental adaptability [12-15], by sacrificing payload due to unchangeable passive compliance. Soft grippers (Fig.1(a1)) have inherent and unparalleled passive compliance [16-19], but also suffer from highly-limited control bandwidth and payload, being also challenging to model and control in real time due to their lack of rigid kinematics. A recent trend of hybrid gripper designs use soft actuators augmented in rigid links [20-25], combining the advantages of soft gripper compliance, and rigid kinematics [24, 25]. However, existing hybrid grippers are mainly focused on high payload and perception, with highly limited dexterity and lateral capability, therefore, cannot cope with the interference force from the lateral direction, nor performing in-hand manipulations[24,25].This could be solved by adding rotation joints to the finger kinematics , but this will increase the mechanism complexity while using rigid-link and motor structure [26], or bring unprecise motion while using soft-approach to simplify the structure [27-31]. Tendon-driven grippers (Fig.1(a3)) combine compliance and high-DOF finger joints using tendon actuation , but the tendon-guiding pulleys will introduce bulkiness and large inertia [32-35].

[1]Shenzhen Key Laboratory of Biomimetic Robotics and Intelligent Systems, Department of Mechanical and Energy Engineering, Southern University of Science and Technology, Shenzhen, China.
[2]Guangdong Provincial Key Laboratory of Human Augmentation and Rehabilitation Robotics in Universities, Southern University of Science and Technology, Shenzhen, China.
[3]Department of Mechanical and Energy Engineering, Southern University of Science and Technology, Shenzhen, China.
[4]Changchun Institute of Optics, Fine Mechanics and Physics, University of Chinese Academy of Sciences, Beijing 100049, China.
#These authors contributed equally to this work.
*Corresponding author, e-mail: wangz@sustech.edu.cn

In this work, we proposed a novel dexterous hybrid gripper design, to enable lateral compliance and manipulation dexterity without losing lightweight and compactness. The gripper comprised of pneumatic-driven soft-robotic joints augmented into rigid structural constraints inspired by the human Metacarpophalangeal (MCP) joint (2-DOF compliant motion) and the Proximal interphalangeal (PIP) joint (1-DOF compliant motion), with enhanced performances on lateral adaptability and compliance, as well as in-hand manipulability. The main contributions of this paper can be summarized as follows:

1) Proposed a novel Pneumatic Synergistic Alignment Joint (PSAJ) inspired by the human MCP joint (Fig.1(b1)), using three independently actuated soft muscles to achieve 2-DOF rotation along pitch and yaw axis, constrained by a rigid universal joint (Fig.1(b2)). Both of the 2-DOF compliance are preserved from the low passive stiffness of soft muscles.

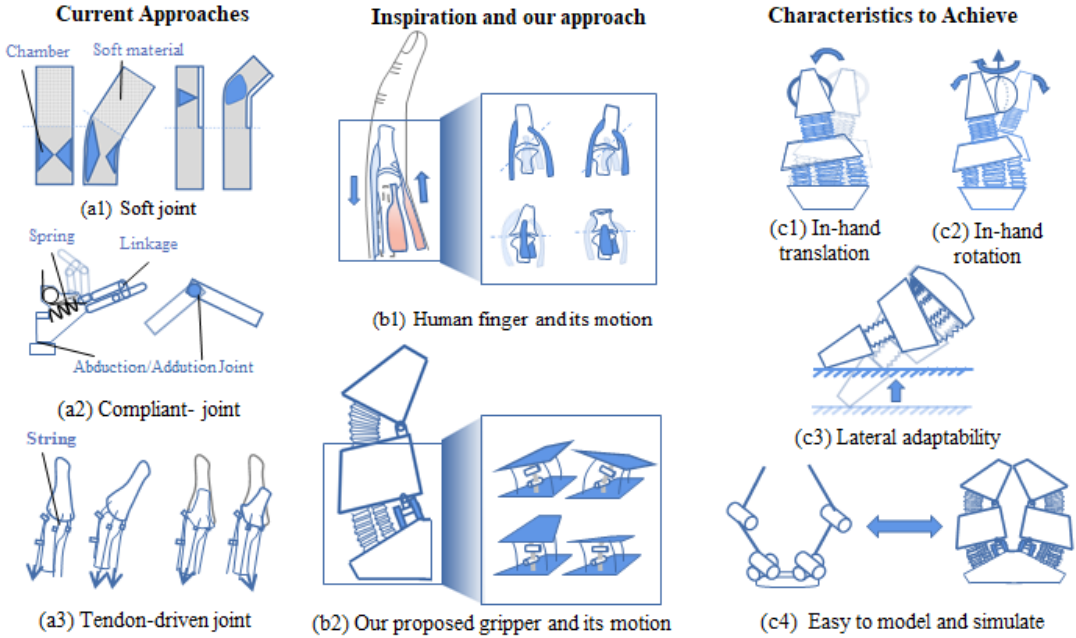

Fig. 1. Concept of the proposed gripper design. (a1) soft hand utilizes a chamber as a joint. (a2) motor-driver gripper achieves abduction/adduction motion by adding rotation joint. (a3) tendon-driven gripper achieves built-in compliance and high DOFs by using string. (b1) inspiration of our work. (b2) model of our proposed gripper and the motion of joint PBSJ. (c) Characteristics with enhanced in-hand manipulability, lateral adaptability, and modeling.

2) Proposed a novel 6-DOF Biorobotic Hybrid Gripper (BHG-6) with PSAJ in each finger, achieving dexterous in-hand rotation and translation of objects, as well as improved lateral payload via passive compliance.

3) Explored a compliance-centric grasping approach of planar soft objects against flat rigid surfaces, where the BHG-6 gripper presses onto the planar object against the rigid surface to form a secure contact and then establish a successful grasp, using the passive compliance of the gripper to avoid large stress. This is validated by picking up a towel from a flat rigid tabletop using the BHG-6 gripper mounted on a dual-arm robot.

The remainder of this article is organized as follows: Sections II demonstrates the inspiration from human MCP joint, together with design and workspace analyzed of our proposed joint and gripper. Section III presents the overall modeling, control and simulation setup of the gripper, followed by the application on towel-pick-up task in section IV. Experimental validations are conducted in Section V, with conclusions in Section VI.

## II. DESIGN OF THE GRIPPER

### A. Inspiration and Design Concept

Human finger joint structure provides us abundant design innovation for robot design. The flexion/abduction/adduction motions of the finger are largely enabled by the MCP joint, with the left-and-right swing of the forefinger realized by the tendons at both ends pulling the two ends of the links on both sides of the adjacent links to move closer or away in the corresponding direction; at the same time, the skeleton between the knuckles provides restrictions on the movement of the fingers. In addition to performing flexion and extension, the dorsal interosseous muscle and palmar interosseous muscle also implement abduction and adduction, see Fig.1(b1). Therefore, the joint could achieve 2-DOF motion within a compact space, with a highly symmetrical and uniform lateral compliance.

Inspired from the MCP joint, we designed a pneumatic soft-rigid hybrid joint that can realize bending and swinging motion (Fig.1(b2)), based on which we designed a 6-DOF symmetrical hybrid gripper. This gripper can be simply modeled and is easy to simulate, also, it has high performances in in-hand translation, rotation and lateral characteristics, see Fig.1(c).

### B. Design and Modeling of PSAJ

The proposed PSAJ joint consists of two rigid links driven by three actuators. To enable abduction/adduction motion, we designed a universal pin hinge as a rigid limiting element to achieve precise motion constraints, as shown in Fig. 2(a). To facilitate the installation of universal pin hinge, there is a certain vertical distance between the two axes of the universal joint.

In order to take into account the flexibility and the manufacturing time of the actuator, we use the blow-molded bellows as soft actuators. The material of the actuator is Ethylene-Vinyl Acetate Copolymer (EVA). Our previous work [29] has proved that under the same shear stress, bellows require low driving force yielding high payload-to-weight ratio as well as an excellent expansion ratio of 2.5:1.

The PSAJ is built around a universal joint with two orthogonal DOFs, corresponding to the abduction/adduction and flexion/extension motion of the human MCP joint, respectively. Following the hybrid design approach with soft actuators augmented into rigid constraints, we use three soft bellows actuators, the minimum number required to actuate two DOFs, and they are all mounted on a common base with the universal joint, such that joint elongation is restricted by the universal joint, only allowing two-way bending (Fig. 2(b)). Here a synergistic muscle alignment was chosen over the antagonistic or orthogonal approaches, in order to achieve both lateral bending (abduction/adduction) while enabling a "power mode" in the flexion direction, where all three muscles could be actuated simultaneously to get the maximum grasping force. This leads to a dedicated grasping mode in control.

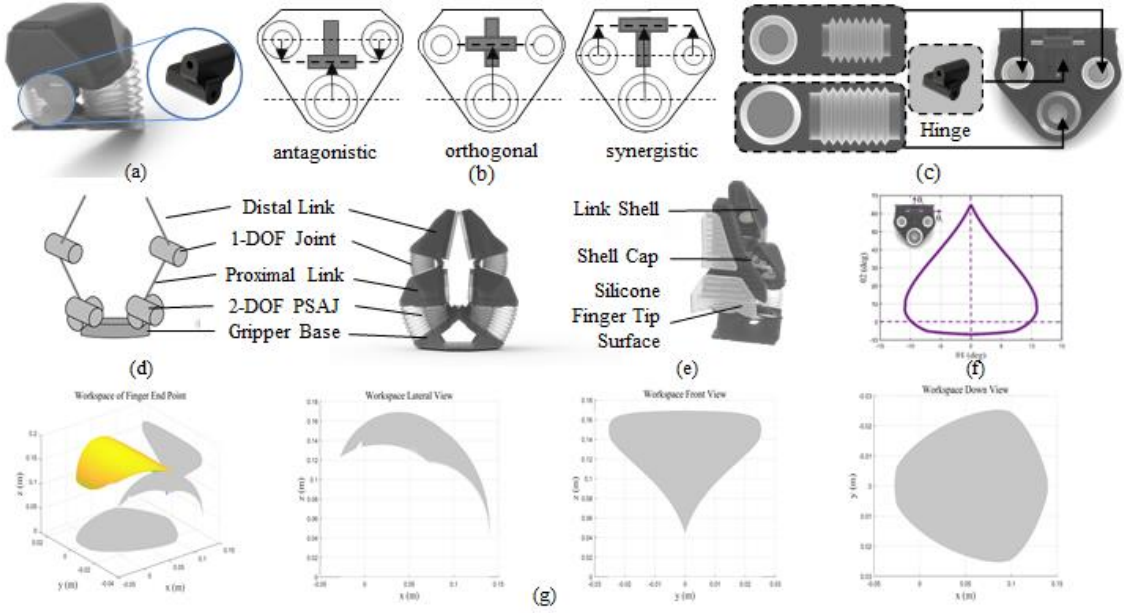

Fig. 2. Details of the PSAJ and BCL-HG. (a) The universal pin hinge. (b) Three available options for actuator distribution. (c) Bellow type. (d) Links and Joints of our proposed gripper. (e) Three layers of gripper. (f) Workspace of PSAJ. (g) Workspace of gripper's end point.

As a result, the joint can realize bending and swing motion under different actuating modes: when there is a difference in left and right actuators, the joint can realize swing motion; when three actuators are simultaneously actuated or only the middle actuator is actuated, flexion is generated, see Fig. 1(b2).

In order to ensure the compactness of the joints, the lateral muscles had smaller diameter than the main muscle, to ensure sufficient flexion force in grasping, as shown in Fig. 2(c). The left-and-right swinging motion is realized by symmetrical small-sized actuators on both sides, and the opening-and-closing motion is realized by the extension and contraction of the large-sized actuator. Detailed size of the two kinds of actuators is shown in Table I. It is worth noting that the dimensions of the two type of bellows are not specifically unique, they will determine other performance indices such as bending range, payload, etc., without losing generality. In this configuration, the workspace analysis of the PSAJ joint is shown in Fig. 2(f).

### C. Design and Modeling of BHG-6

The proposed BHG-6 design adopts a hybrid symmetrical two-finger design. Similar to the human hand, the overall gripper consists of three layers:

**Layer 1-Rigid links**. The 3D-printed components consist of 3 different types: distal link, proximal link and gripper base. Distal link and proximal link are joined by a general pin hinge, which is restricted from the lateral relative movement of the knuckles of the gripper's fingers during grasping; proximal link and gripper base are joined by universal pin hinge.

TABLE I
GEOMETRY OF BELLOWS

| | | |
|---|---|---|
| $l_1$ | Length of large bellow's elbow-cover area | 70mm |
| $l_2$ | Length of large bellow | 44mm |
| $l_1'$ | Length of small bellow's elbow-cover area | 50mm |
| $l_2'$ | Length of small bellow | 39mm |
| $d_1$ | Diameter of big bellow | 30.5mm |
| $d_2$ | Diameter of big bellow's elbow-cover area | 23mm |
| $d_1'$ | Diameter of small bellow | 37.56mm |
| $d_2'$ | Diameter of small bellow's elbow-cover area | 32.37mm |

**Layer 2-Soft Actuators.** The gripper consists of 4 large-sized bellows and 4 small-sized bellows, the presence of which provide good compliance for gripper, as shown in Fig. 2(d). Bellow 0 and 4 drive finger-tip; bellow 1,2,3/4,5,6 are components of PSAJ, which drives the finger root.

**Layer 3-Contact surface and inner piping.** The contact surface of each finger is molded from silicone rubber (Ecoflex-0030) with symmetric texture in both directions, see Fig. 2(e). When grasping, the flexible contact surface can increase the contact area and improve grasping stability. The finger links are designed to be hollow for weight reduction and passing pipes.

Finally, the working space sanalysis of the gripper is shown in the Fig. 2(g), with the tip joint movement range of the fingertip measured experimentally.

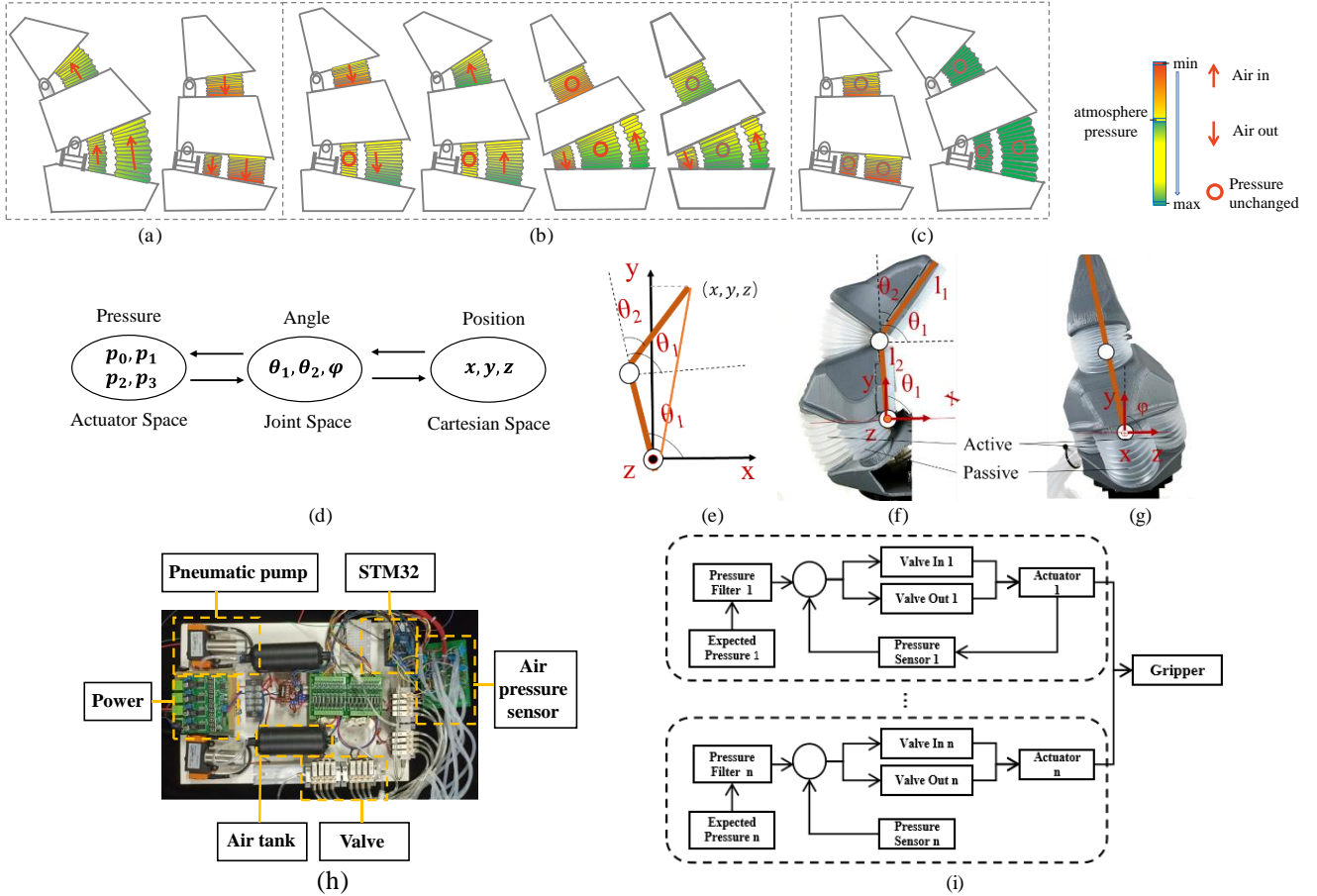

Fig. 3. Low-level Control of Gripper. (a)-(c)Control mode of the gripper and the mapping between bellow's pressure with color. (a) Mode 1 ("Power mode"). (b) Mode 2 ("Ab/Ad mode").(c) Mode 3 ("Holding mode"). (d) The mapping between three spaces which defines the kinematics of hybrid bellow gripper. (e) Kinematics schematic when gripper rotates in x-y plane. (f) – (g)Geometric method to solve the inverse kinematics. (h) The real control device of the control system. (i) The structure of the pressure control system.

## III. KINEMATICS AND CONTROL OF THE GRIPPER

### A. Low-level Control of the Gripper

#### a. Control Modes

The need of different gripper performance like grasping force or control simplicity various according to different task. Facing with the various need, the synergistic aligned PSAJ provided us possibilities to deal with the need by only changing the actuation logic. Here, we divide the actuation logic into three working modes to deal with different application need:

**Mode 1 ("Power mode").** Power grasping needs large grasping force, this mode is designed to provide large grasping force. Under synergistic alignment of actuators, the pressure of at the end of each actuator will turn into torque at direction of closing gripper. Utilizing this characteristic, in this mode we inflate ALL of the actuators in closing the gripper to maximize the output grasping force of our gripper, see Fig. 3(a).

**Mode 2 ("Ab/Ad mode").** The three muscles used in the PSAJ could independently control the two DOFs to perform very dexterous motions combining abduction/adduction and flexion/extension. To simplified the control law, we decoupled the angle mapping to actuator pressure. The inflation and deflation of actuator at mid axis response for the open-close angle of the PSAJ. The pressure difference of actuator 1 and 3 results in lateral rotation, see Fig.3(b).

**Mode 3 ("Holding mode").** Besides two modes above, we also proposed a mode to stabilize the finger's position. In this mode, valves are shut to lock down air in the actuators, then the valve connecting actuators are closed to lock down the air in actuator. When external force removed, the finger will recover to original position before applying external force, which will stabilize the grasping, see Fig. 3(c).

#### b. Kinematics and Pressure Control Mapping

The detailed control flow of configuration of the finger mentioned in mode 2 is shown in Fig.3(d), We firstly set up relationship between end point of gripper $\{x, y, z\}$, (cartesian space position of the fingertip), and joint angle $\{\theta_0, \theta_1, \varphi\}$ (representing the rotation angle of top joint in x-y plane, root joint in $x$-$y$ plane, root joint in $y$-$z$ plane, respectively) .Then relationship between joint angle and actuator pressure $\{p_0, p_1, p_2, p_3\}$ (pressure of 4 actuators, in order from top to root and from left to right.) is studied, see Fig.3(e).

Benefiting from adopting a hybrid structure, compared to continuum soft grippers which need material parameters and simulation to model the configuration, our gripper tip can be model through the geometry of links and axis.

From $x$-$y$ plane, as shown in Fig. 3(f), the position of the fingertip can be described as

$$\mathbf{P_e} = \begin{pmatrix} l_1\cos(\theta_1-\theta_2)+l_2\cos\theta_1 \\ l_1\sin(\theta_1-\theta_2)\cos\varphi+l_2\sin\theta_1 \\ 0 \end{pmatrix} \quad (5)$$

When $P_e$ is rotated around $x$-axis, as shown in Fig.3(g), the position becomes

$$\mathbf{P_e'} = \mathbf{R_x}\mathbf{P_e} = \begin{pmatrix} l_1\cos(\theta_1-\theta_2)+l_2\cos\theta_1 \\ l_1\sin(\theta_1-\theta_2)\cos\varphi+l_2\sin\theta_1\cos\varphi \\ l_1\sin(\theta_1-\theta_2)\sin\varphi+l_2\sin\theta_1\sin\varphi \end{pmatrix} \quad (6)$$

For joint angle control, we linearized the mapping between joint angle and pressure method to control joint angle following a similar approach reported in previous work [24, 25].

Combining with decouple method mentioned in mode 2, the rotate angle of the tip joint $\theta_1$ is regarded as a linear function of pressure at actuator 0, which can be presented as $\theta_1 = k_0 p_0 + b_0$. Similarly, open-close angle $\theta_2 = k_2 p_2 + b_2$, lateral rotational angle $\varphi = k_3\Delta p_3 + b_3$, where $\Delta p_3$ denotes the difference between actuators 1 and 3. Substituting the linear equations in (6), we have,

$$\mathbf{P_e'} = \begin{pmatrix} l_1 c(k_0 p_0 - k_2 p_2 + b) \\ + l_1 c(k_0 p_0 + b_0) \\ \\ l_1 s(k_0 p_0 - k_2 p_2 + b)c(k_3\Delta p_3 + b_3) \\ + l_2 s(k_0 p_0 + b_0)c(k_3\Delta p_3 + b_3) \\ \\ l_1 s(k_0 p_0 - k_2 p_2 + b)s(k_3\Delta p_3 + b_3) \\ + l_2 s(k_0 p_0 + b_0)s(k_3\Delta p_3 + b_3) \end{pmatrix} \quad (7)$$

The model of joint angle and pressure is validated in experiments.

#### c. Low-level Control Workflow and hardware

To validate our proposed control method, we set up a pneumatic control board, as shown in Fig. 3(h). The pneumatic system of the control board consists of two pumps (KVP8S, Kamoer Inc.), two air tanks and solenoid valves (OST Inc.) and an MCU of STM32F767 (ST Microelectrons Inc.) is used as the control unit of the system, and pressure sensors (SSCDANN060PAAA5, Honeywell Inc.) are used to provide pressure feedback. The board is set to receive pressure command or valve action command, the control algorithm of pressure is shown in Fig. 3(i), which is also adopted in our previous work [6], [28].

### B. High-level Control System of the Gripper

To validate the dexterity and passive-adaptability of our proposed gripper and showcase their benefit to the performance of ADL, we developed a proprietary high-level control system including a dual-arm robot and RGBD vision to provide perception and to move the gripper for ADL tasking of soft, deformable and thin objects appressed on rigid surfaces, e.g. a towel spread out on a table, requires compliance and passive adaptability of the gripper. To validate the advantages of our gripper on such tasks, we present a visual-based planning process utilizing our proposed gripper combine with the high-level control hardware.

#### a. Hardware Specification

The hardware of high-level control system is shown in Fig. 4(a). Grippers are installed as the end effector of dual-arm robot with 6-DOF for each arm. (KV220, Santifico). An RGB-D camera (Azure Kinect, Microsoft) was installed at the head of dual arm robot for object recognition and targeting.

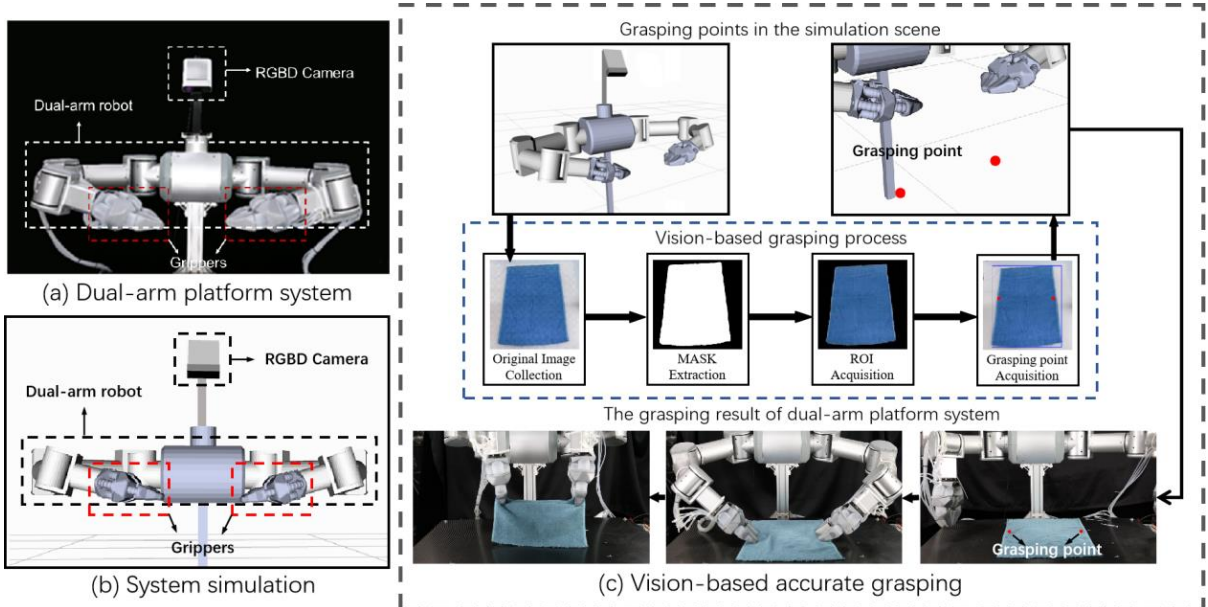

Fig.4. The real-world platform system. (a) The system includes a dual-arm robot, two grippers and a camera, which obtain (b) and (c). (b) Planning-based Fuzzy Grasping mode. (a) The grippers arrive at fuzzy specified position, and the grippers grab the towel and lift it. (c) The process of vision-based Accurate Grasping mode.

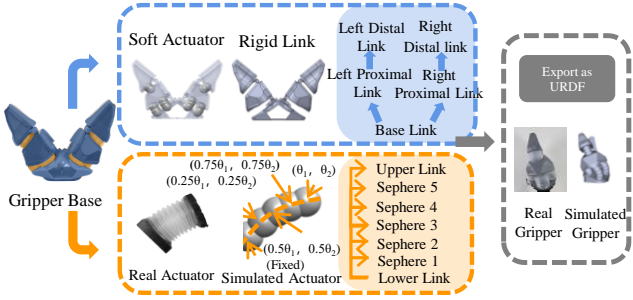

Fig. 5. Setup workflow of gripper model for simulation and visualization. To compatible with current available robot simulation and visualization software like Rviz, the whole model, including the bellow modeling, are achieve using multi-rigid body method. The rigid structure part, marked using blue color, can be modeled naturally for its rigid structure. The soft actuator is separated in several parts. Each part is represented using a sphere. The final performance of our simulation model is shown in grey box.

### b. Simulation Model Setup of the Gripper

To integrate the gripper into high-level robot system like Rviz or ROS [34] for robot arm to plan and avoid obstacle in unstructured environment, we build the simulation model for gripper in hybrid structure. Different from soft gripper which is difficult to simulate with their continuum body, hybrid grippers are easy to construct such simulation model for their rigid-link structures can be modeled similarly to rigid-link grippers.

The model setup work-flow of the gripper is shown in Fig.5. The whole setup process can be separated for the rigid links and soft actuators, respectively. For rigid links, two fingers are modeled as two serial links connected to the same base link. The actuators were equally divided into five segments, each represented using a sphere. The bottom sphere is fixed with the bottom link while the top sphere is fixed with the upper link. The sphere between is treated to rotate among axis with proportional angle. Simulation model with dual-arm system is shown in Fig. 4(b).

### c. Visual-based Grasping Point Detection

Compare to traditional towel grasping detection which need to detect grasping point by detecting spatial edge, acquisition of the grasping midpoints is done through a five-step 2.5D grasping point acquisition approach, the edge detection of towel edge is determined using only one 2D image at a time. Detailed process include:

(1) Collect the original image;
(2) Extract the MASK;
(3) Segment region of interest (ROI);
(4) Detect 2D grasping points;
(5) Calculate the actual grasping point position.

The vision-based grasping process is shown in Fig. 4(c). Due to the different color of the object and background, blue color of towel is used as a feature to separate the ROI, and the MASK range of the blue area is [100,100,50]-[130,255,255]. To obtain the minimum enclosing rectangle of ROI, the edge detection of ROI is carried out by Canny operator. The detection of 2D grasping points is the midpoints of the minimum enclosing rectangle.

After acquiring the coordinate of grasping point, the coordinate is transformed from image coordinate system to world coordinate system for the following planning and grasping.

### d. Passive Adaptability to Grasp Against Rigid Surface

Grasping towel against rigid surface need compliance of gripper to reduce the impact force applied to the gripper cause by rigid surface in the process of grasping. Besides, the depth error (about 11 mm according to [35]) of our vision system result in inaccuracy positioning in both transversal, longitudinal and vertical direction, where the transversal and longitudinal positioning error are avoided by setting grasping point more inside the towel. However, the vertical error is not able to avoid by regulating grasping point in visual-based process. To mitigate the error as well as producing stable grasping against rigid surface, we proposed a 'poke and pinch' process to grasp the towel.

The whole process is shown in Fig.4(c). Before grasping, the end effector points are set to be above of the detected grasping point, with gripper inclined to allow finger passively-deform in lateral direction at poking process.

Under Mode 2, the gripper was set to be open in pinching configuration. Reach target of the end point is set beneath the detected position for 20-30 mm, in order to over-press and deflect the gripper onto the surface. Benefiting from the lateral compliance, the PSAJ gripper will be adapt passively, establishing contact with the towel while remaining open. Using mode 2, the gripper closes the fingers to pinch the towel, then switch to mode 3, to hold the grasping pressure while the gripper picks the towel up. The error tolerance under different approaching angles were studied through experiments and reported in the next chapter.

## IV. EXPERIMENTAL VALIDATION

### A. Gripper characteristic validation

The comparisons between the proposed PSAJ gripper and previous works are shown in Table IV. Compared to previous work, our proposed hand is much lighter with a high payload as well as high dexterity resulting from its higher DOF per finger.

#### a. Single joint characteristics

The rotating angle of a single joint was validated first, as shown in Fig. 6(a). We placed the grippers before a calibration board and obtain the image of them, then corrected the distortion of image. A visual protractor was used to measure the rotating angle of joints. Then, based on acquired data, the relations between angles and joints proposed in Eq. (7) can be obtained and validated. Results of curve fitting are presented in Fig.6(b)-(d), with RMSE of $5.151°$, $1.202°$ and $2.122°$, respectively.

#### b. Gripper payload test

The maximum load in different directions of the gripper was also validated, as shown in Fig. 7. The test apparatus includes a tension meter capable of storing the peak tension. Connected to the tension meter is a small ball caught in the paws. At the same time, a rigid limit structure is installed on both sides of the hand to limit the lateral movement of the hand. By adding or removing this limit structure, we can compare the advantages of adding lateral degrees of freedom.

TABLE II
COMPARISON OF THE PROPOSED GRIPPER AND PREVIOUS WORKS

| | BCL-4 [19] | BCL-6 [24] | BCL-13 [28] | Proprioceptive hybrid gripper [25] | **Our proposed gripper** |
|---|---|---|---|---|---|
| *Fingers* | 2 | 3 | 4 | 2 | **2** |
| *DOFs* | 4 | 6 | 13 | 3 | **6** |
| *Self-weight* | 0.2kg | 0.4kg | 1.27kg | 3.2kg | **0.39kg** |
| *Grasping Force* | Joint:50N, Distal:12N | 40N | 9.6N | Joint:564.5N, Proximal:302.4N, Distal:52.1N | **Proximal:84.04N, Joint:35.73N, Distal:19.87N,** |

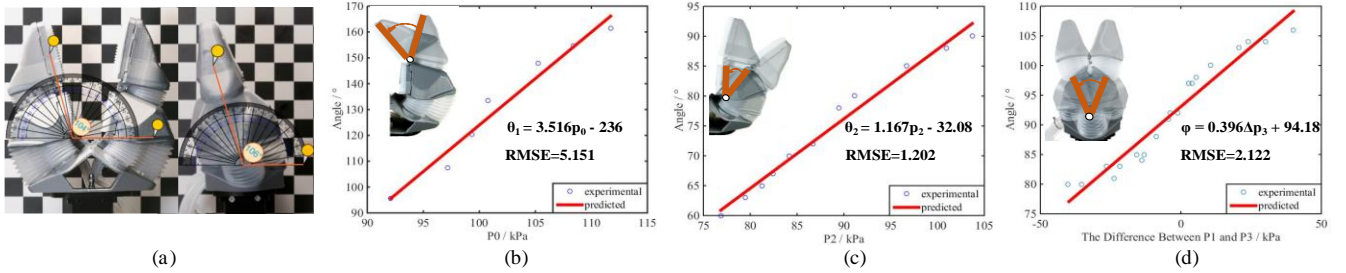

Fig. 6. Angle measurement method and validation of each joint linear characteristic between rotate angle and pressure. (a) Measurement method of rotation angle. (b) The tip joint in x-y plane. (c) The root joint in x-y plane. (d) The root joint in y-z plane.

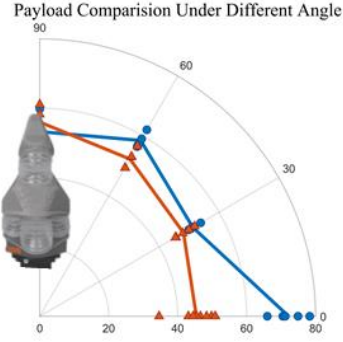

Fig. 7. Load test experiments. (a) Load-test experiments with limiting the lateral movement. (b) Load-test experiments without limiting the lateral movement.

The load plots in Fig. 7 show the gripper was able to carry load as 25 times heavy as its self-weight; In contrast, the load capacity was reduced to less than 50N when the lateral DOF was deleted, 53.5N lower than the former situation. This was largely due to the lateral compliance enabled by the PSAJ joint, allowing the gripper to deflect towards the force loading direction, significantly improving the grasping envelop. As grasping is most secure with the PIP and MCP joints both wrapping around the ball, generating maximum friction on the silicone soft contact layer against external pulling forces, this passive adaptation resulted in substantially higher pull-out forces, of 87.3% over the case when lateral compliance was restricted by the limiter.

#### c. Tracking performance

Three different commands were used in driving the actuators to assess the tracking performance of the control system.

**Group1:** the left side gripper and right-side gripper are simultaneously opening and closing.

**Group2:** the left side gripper and right-side gripper are simultaneously swinging to the same direction.

**Group3:** the left side gripper and right-side gripper are simultaneously swinging to the different direction.

The interval of each group starting at 1.3s, reduces 0.08s each time after a wave, and end while the interval smaller than 0.2s.

As can be seen from Fig. 8, the pressure could reach the reference at least once during the cycle for frequencies up to 5 Hz.

### B. Gripper performance validation

#### a. In-hand manipulation

Experiments were carried out to test the gripper's grasping adaptability, as shown in Fig. 8. We prepared three grabbing items with different shapes. For smaller-sized wires, the gripper uses fingertips to grab; for medium-sized glue bottle, the gripper uses the middle of the finger to grab; for larger-sized beverage bottle, the gripper uses the finger base to grab it, see Fig. 9(a)-(c). In Fig. 9(d) - (e), the manipulability is validated using a 10cm ball. Rotation and simply translation of the ball is tested. In Fig. 9(f), We also tested small and difficult-to-operate objects such as pens, and performed tasks such as turning the pen with the gripper.

#### b. Lateral DOF performance

As we all know, the ordinary gripper needs to tilt the cup with the help of the robotic arm when it is performing the task of pouring water, but when our proposed gripper is performing this task, the robotic arm does not need to perform any operation, the task only needs to rely on the side swing of the gripper to pour out the water, which greatly reflects the superiority of the hand we proposed, see Fig. 10.

### C. Dual-gripper performance test

#### a. Towel folding performance

Fig. 10 shows a successful run of the pipeline applied to a towel. Once a towel is grasped and lifted up with the help of robotic arms (Fig. 11 (a)-(b)), the grippers start its folding and delivering process. The whole procedure is performed by two grippers until the left gripper move away with holding the whole towel (Fig.11 (c)-(d)). Due to the flexibility and wide working space, the towel folding and delivering work can still be performed purely via in-hand manipulation of the grippers, without moving the manipulator arms. Demonstration of the towel-passing procedure by in-hand manipulation is included in the supplemented video.

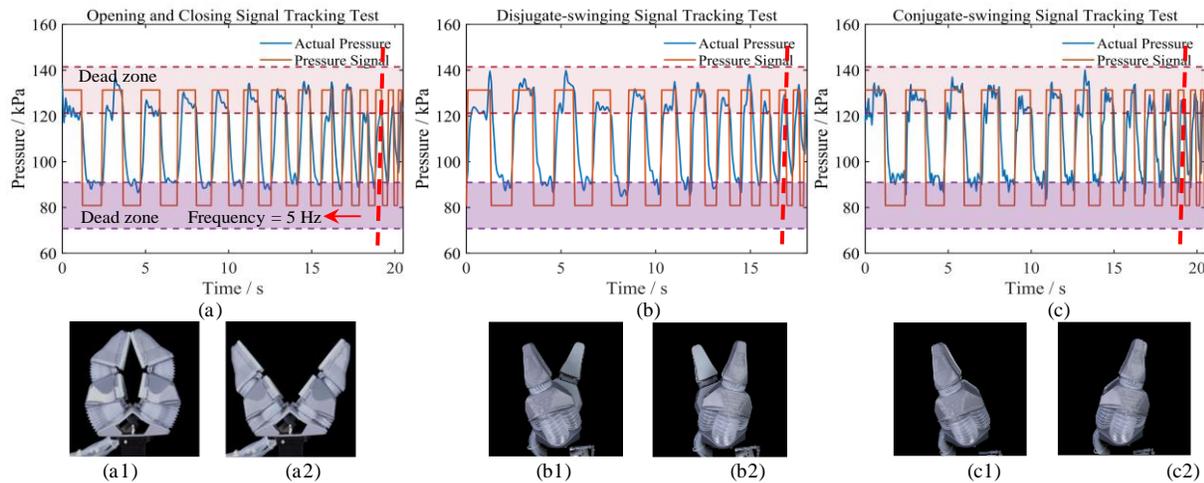

Fig. 8. Tracking performance test. (a)Tracking performance when gripper opening and closing in continuous circulation. (b) Tracking performance when gripper circulate in the same direction simultaneously. (c) Tracking performance when gripper continuously rotate in opposite directions.

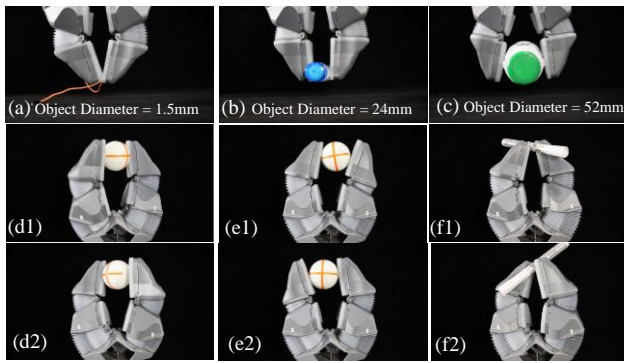

Fig. 9. In-hand manipulation experiments. (a) Grasping a wire. (b) Grasping a glue bottle. (c) Grasping a beverage bottle. (d) In-hand rotation of rolling a ball. (e) In-hand translation of rolling a ball. (f) In-hand rotation of rolling a pen.

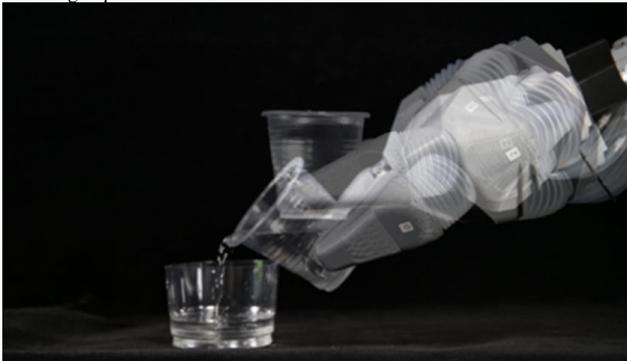

Fig. 10. Pouring water procedure. It is important to note that there is no wrist movement in this figure.

### b. Lateral compliance test

As shown in Fig. 12, the rectangular object to be grasped is composed of two independent aluminum profiles of 2cm*2cm, and a 6-DOF force sensor. The force sensor is responsible for detecting the relationship between the force and moment transmitted in the rod. In the experiment, the right arm is only used for clamping, and the left arm is clamped and moved 1cm outward to simulate the relative position error caused by the robot arm error or poor coordination during the collaborative grasping process.

Depending on the enabled lateral compliance on the pulling and receiving ends, the experiments were divided into four different modes, as shown in Fig. 12 (b)-(e): lateral compliance enabled on both ends; lateral compliance enabled only on the pulling end; lateral compliance enabled only on the recipient end; no lateral compliance on either end. The results are shown in Fig. 12 (a). For the group without lateral freedom, the stress of this group is the largest and the rising slope is the largest. For the mode with single lateral freedom, the stress is only that there is no lateral freedom. For the group with two degrees of freedom, the stress is only 47.2% of the one-sided degrees of freedom and 10.6% of the group without lateral degrees of freedom. Enabling lateral compliance, therefore, could effectively and substantially reduce closed-loop stress and enabling the proposed PSAJ gripper to handle rigid objects bimanually without risking of any damaging or instability.

### c. Lateral compliance for object picking from rigid surfaces

Finally, the capability of the proposed PSAJ gripper in picking up flat objects against rigid surfaces was assessed. The gripper utilized its inherent passive compliance as a means to adapt to the misalignment between the manipulator end effector and the rigid surface it is pressed against. Therefore, it could pick up flat objects from rigid surfaces without damaging either the surface, or itself, without requiring a deformable surface as for conventional rigid robotic grippers.

To validate the passive compliance performance, the experiment was divided into five conditions, where the gripper approached the rigid surface from different angles (0,15, 30, 45, and 60 degrees, respectively). Each experiment was repeated for several times, with each time the vertical error (how much distance the gripper is being pressed into the rigid surface) increased from 0 to 10 mm at 1 mm intervals.

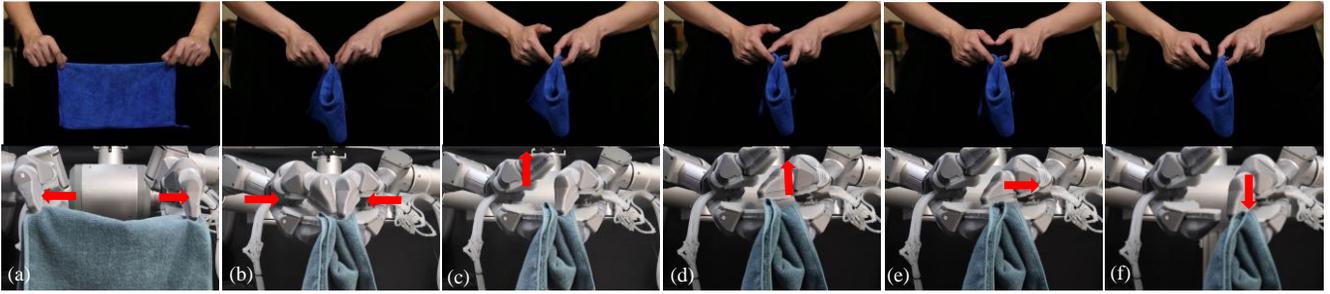

Fig. 11. Complete pipeline of folding a towel. (a) The towel was successfully picked up by grippers. (b) Fold the towel with two grippers. (c) Loose the towel from the upper left gripper. (d) Loose the towel from the upper right gripper. (e) Move the towel on the right side to the left side with the upper left gripper. (f) Grab the towel with right gripper and move away.

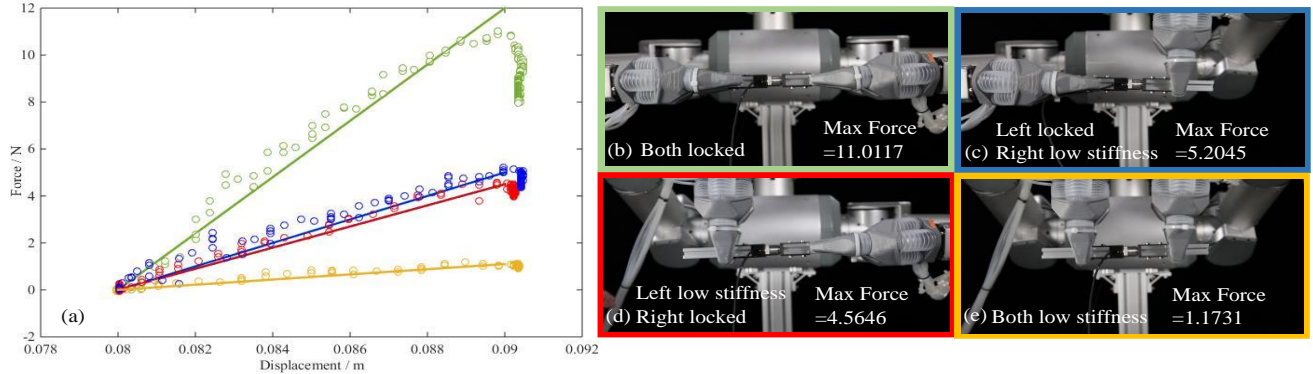

Fig. 12. Dual-gripper lateral compliance experiments. (a) The relationship between pulling force and pulling displacement of each experiment. (b)-(e) Pulling a aluminum alloy rod with two arms.

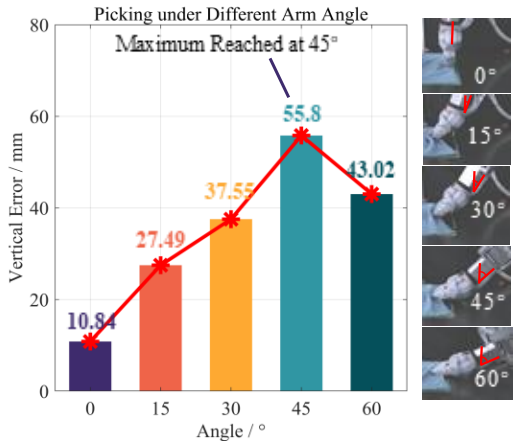

Fig. 13. Lateral compliance experiments. The gripper grasps the towel at 15°, 30°, 45°, 60° and 90° inclines, respectively.

As shown in Fig. 13, results from the five experiments, the gripper reached down and grabbed the towel at different approaching angles, the maximum tolerated error was recorded, where the gripper could still pick up the towel successfully, despite the vertical error in manipulator position. It could be observed from the results that the fault tolerance is greatest when the gripper is at an approaching angle of 45°, and decreases when the angle both increases and decreases. This is most likely due to the mechanistic constraint of the rigid links, where the soft actuator's compliance will be hindered and even eliminated when being fully compressed between two mounting rigid parts. At the same time, the results in Fig. 13 highlighted the excellent passive adaptability of the PSAJ gripper against environmental interaction uncertainties. Across the five tested approaching angles, at least 1 mm of error tolerance was observed, enabling the gripper to effectively picking up a very thin and soft object (a towel, in this case), from a rigid desk.

## V. CONCLUSIONS AND FUTURE WORK

In this paper, a new dexterous 2-finger hybrid gripper design was proposed with 6 DOFs actuated by 8 independent pneumatic soft muscles, augmented into rigid constraining structures. Inspired by the human MCP and PIP joints, the gripper had both flexion/extension and abduction/adduction enabled on the proximal joint, and flexion/extension on the distal joint. It was shown that adding the MCP complexity with a synergistic muscle configuration was effective in enabling a range of novel features in the hybrid gripper, including in-hand manipulation, lateral passive compliance, as well as new control modes. A prototype gripper was fabricated and tested on our proprietary dual-arm robot platform with vision guided grasping. With very lightweight pneumatic bellows soft actuators, the gripper could grasp objects over 25 times its own weight with lateral compliance.

Kinematic modeling was performed to describe the motion of the gripper. A multimode method is proposed to enable the gripper to have the ability to grasp and operate a wide range of objects and complete dexterous tasks. The load test showed the lateral DOF provides 59.1 N more than without lateral DOF. For dual-arm manipulation capability, the grippers are proved for complex tasks like folding towels, and the lateral stiffness can also reduce the stress by 70%. Utilizing the gripper's lateral compliance, a list of interactive features were achieved, from reducing closed-loop stress, to adapting to vertical error during thin object picking against rigid surfaces. The PSAJ gripper, completed with modeling and the overall dual-arm system demonstration, offers a high-performance alternative to compliant gripper design.

Future works will focus on mechanical design refinements; higher level control iterations; as well as adding sensing modalities to improve interactive intelligence. This work paves the way to soft-rigid hybrid grippers with a strategically balanced blend of simplicity, adaptability, and dexterity.

## Acknowledgement

The authors want like to thank Mr Kehan Zou for his help in setting up hardware of control device, and Ms Yu Cheng for casting the silicone layer of the gripper.

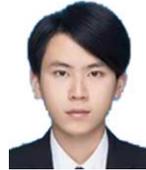
**Wenpei Zhu** received his B.S degree in Mechanical Engineering from the Hefei university of science and technology in 2018. He is currently a master student majoring in Electronic science and Engineering in the Department of Mechanical and Energy Engineering, Southern University of Science and Technology, Shenzhen, China. He is currently working in the BioRobotics and Control Lab. His research interests include control and modelling of soft robotics and dual-arm manipulator.

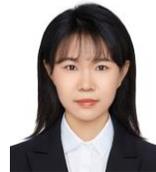
**Chenghua Lu** received the B.S. degree in Mechanical Engineering from the Northeastern University and M.S. degree in Mechanical Manufacturing and Automation from the University of Chinese Academy of Sciences. She has been a visiting graduate student in the BioRobotics and Control Lab at the Southern University of Science and Technology in 2020. Her research interests focus on soft hand design, control algorithm optimization and cooperative robot control.

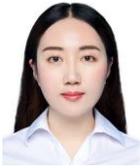

**Qule Zheng** received the B.S. degree in Electronic and Information Engineering from the Northwest A&F University in 2017, and M.S. degree in Aerospace Engineering from the Xidian University in 2020. She is currently a Ph.D. student majoring in Mechanics in the Department of Mechanical and Electrical Engineering, Harbin Institute of Technology, Shenzhen, China. She is currently working in the BioRobotics and Control Lab. Her research interests include digital image processing and computer vision.

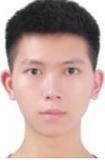

**Zhonggui Fang** received the B.S. degree in Mechanical Design Manufacturing and Automation from the Guangdong University of Technology in 2019, He is currently a postgraduate student pursuing his M.S. degree major in Mechanics in the Department of Mechanical Engineering and Energy Engineering, Southern University of Science and Technology, Shenzhen, China. His research interests include soft robotics for medical application and wearable robots.

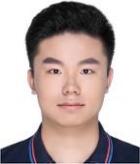

**Haichuan Che** is currently an undergraduate student majoring in Mechanical Engineering in the Department of Mechanical and Engineering, Southern University of Science and Technology, Shenzhen, China. He is currently working in the BioRobotics and Control Lab. His research interests include soft robotics and robot arm manipulation.

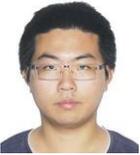

**Kailuan Tang** received the B.S. degree in Communication Engineering from Southern University of Science and Technology, Shenzhen, China, in 2017. He is currently Ph.D. student majoring in Mechanics in the Department of Mechanical and Electrical Engineering, Harbin Institute of Technology, Shenzhen, China. He is currently working in the BioRobotics and Control Lab. His research interests include underwater robots, biomimetic control and soft robots.

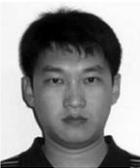

**Mingchao Zhu** received the B.S. degree in automation from Jilin University, Changchun, China in 2003, and the Ph.D. degree in control theory and control engineering from Jilin University in 2009. He is currently a Researcher with the Changchun Institute of Optics, Fine Mechanics and Physics, Chinese Academy of Sciences, Changchun. He is a currently a Master supervisor in University of Chinese Academy of Sciences. His research interests include kinematics, dynamics and robot control.

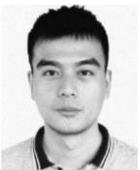

**Sicong Liu** received the B.S.(Hons.) degree in Mechanical Design Manufacturing and Automation from the Harbin Institute of Technology (HIT), Harbin, China, in 2009, the M.S. degree in mechanical design and theory from HIT in 2011, and the Ph.D. degree in robotics and engineering mechanics from Nanyang Technological University, Singapore, in 2015. In 2016, He was a Postdoctoral Research Fellow with Nanyang Technological University, Singapore. He was a mechanical Engineer in DJI in 2017. Then, He was a Senior Mechanical Engineer in UBTECH in 2019. He is currently a Research Assistant Professor with the Department of Mechanical and Energy Engineering and the Institute of Robotics, Southern University of Science and Technology, Shenzhen, China. His research interests include deployable structures inspired by origami and soft robotics.

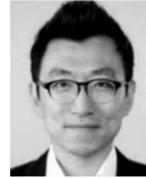

**ZHENG WANG** (S'06_M'10_SM'16) received the B.Sc. degree (Hons.) in automatic control from Tsinghua University, Beijing, China, in 2004, the M.Sc. degree (Hons.) in control systems from the Imperial College London, London U.K., in 2005, and the Ph.D. degree (Hons.) in electrical engineering from Technische Universitaet Muenchen, Munich, Germany, in 2010. He was a Postdoctoral Research Fellow with Nanyang Technological University, Singapore, from 2010 to 2013, and a Postdoctoral Fellow with the School of Engineering and Applied Sciences and the Wyss Institute of Bioinspired Engineering, Harvard University, in 2013 and 2014, respectively. Since July 2014, he has been an Assistant Professor with the Department of Mechanical Engineering, The University of Hong Kong, Hong Kong. He has been a professor in robotics with the Department of Mechanical and Energy Engineering, Southern University of Science and Technology, Shenzhen, China, since February 2019. His research interests include haptics human-robot interaction, endoscopic surgical robot, underwater robots, and soft robotics